\documentclass[conference]{IEEEtran}

\usepackage[utf8]{inputenc}

\usepackage{url}
\usepackage{amsmath}
\usepackage{graphicx}
\usepackage{booktabs}
\usepackage{multirow}
\usepackage[hidelinks]{hyperref} 
\usepackage{algorithm}
\usepackage{algorithmic}
\usepackage{amsmath}
\usepackage{amssymb}

\title{Smart Traffic Signals: Comparing MARL and Fixed-Time Strategies}

\author{
    \IEEEauthorblockN{Saahil Mahato}
    \IEEEauthorblockA{
        Keele University\\
        Suryabinayak, Bhaktapur, Nepal\\
        mahatosaahil4@gmail.com
    }
}

\begin{document}

\maketitle

\begin{abstract}
Urban traffic congestion, particularly at intersections, significantly affects travel time, fuel consumption, and emissions. Traditional fixed-time signal control systems often lack the adaptability to effectively manage dynamic traffic patterns. This study explores the application of multi-agent reinforcement learning (MARL) to optimize traffic signal coordination across multiple intersections within a simulated environment. A simulation was developed to model a network of interconnected intersections with randomly generated vehicle flows to reflect realistic traffic variability. A decentralized MARL controller was implemented in which each traffic signal operates as an autonomous agent, making decisions based on local observations and information from neighboring agents. Performance was evaluated against a baseline fixed-time controller using metrics such as average vehicle wait time and overall throughput. The MARL approach demonstrated statistically significant improvements, including reduced average waiting times and improved throughput. These findings suggest that MARL-based dynamic control strategies hold substantial promise to improve urban traffic management efficiency. More research is recommended to address the challenges of scalability and real-world implementation.
\end{abstract}


\section{Introduction}

Urban traffic congestion is a significant challenge faced by cities worldwide, resulting in longer travel times, higher fuel consumption, and increased environmental pollution \cite{pwc2019urban}. Traditional traffic signal control methods, such as fixed-time control, operate on predetermined schedules that do not adapt to real-time traffic conditions \cite{nacto2025fixed}. Although these systems are easy to implement and maintain, they often do not perform well in dynamic environments characterized by unpredictable traffic patterns \cite{nacto2025fixed}. To overcome these limitations, adaptive traffic control strategies have been developed utilizing real-time data to adjust signal timings for optimal traffic flow \cite{adaptiveTraffic2011}. Recent advances in artificial intelligence, particularly in Reinforcement Learning (RL), have introduced Multi-Agent Reinforcement Learning (MARL), wherein each traffic signal acts as an autonomous agent capable of learning optimal control policies through environmental interactions \cite{li2022multiagent}. This study aims to conduct a comparative analysis of the performance of a traditional fixed-time controller versus a MARL-based controller in a simulated multi-intersection traffic environment, using computer simulation to model a network with randomly spawned vehicles that emulate real-world traffic variability. The research specifically examines the effectiveness of MARL in reducing average vehicle travel time, minimizing queue lengths, and improving overall traffic throughput compared to fixed-time control systems.

\section{Research Objective}

The primary objective of this research is to conduct a comparative analysis between a fixed-time traffic signal controller and a multi-agent reinforcement learning (MARL)-based controller within a simulated multi-intersection traffic network. The study aims to evaluate the performance of each control strategy by simulating realistic traffic flow scenarios, systematically recording key performance metrics such as average waiting time and throughput. Through statistical comparison of the results, the research seeks to determine the relative effectiveness of each approach and assess the significance of any observed differences, thus providing empirical insight into the potential advantages of adaptive, learning-based traffic control systems over traditional fixed-time methods.

\section{Related Work}

Optimization of traffic signal control has been a central theme of research for decades, evolving significantly from traditional static timing strategies to more sophisticated intelligent systems. This section critically evaluates the progression of traffic signal control methodologies, elucidating the limitations of conventional approaches while underscoring the advancements introduced by artificial intelligence, with a particular focus on Multi-agent Reinforcement Learning (MARL).

\subsection{Fixed-Time and Actuated Signal Control}

Conventional traffic signal control methods, particularly fixed-time control, operate on the basis of established schedules that do not adjust to real-time traffic conditions \cite{nacto2025fixed}. Although these systems are relatively straightforward to implement, their static nature often results in inefficiencies in dynamic traffic environments \cite{web25}. In contrast to this, actuated signal control introduced a degree of responsiveness by adjusting the signal phases based on data obtained from sensors \cite{nacto2025fixed}. However, despite this advancement, such systems still lack the requisite adaptability to effectively manage the complexities inherent in urban traffic scenarios \cite{web26}.

\subsection{Adaptive Traffic Control Systems (ATCS)}

To address the deficiencies associated with static control methods, Adaptive Traffic Control Systems (ATCS) emerged as a progressive alternative. Prominent examples of ATCS include the Split Cycle Offset Optimization Technique (SCOOT) and the Sydney Coordinated Adaptive Traffic System (SCATS). SCOOT uses real-time traffic flow data to adaptively adjust signal timings, demonstrating performance enhancements of approximately 15\% relative to fixed-time systems \cite{ijlemrSCOOT}. SCATS, which has been successfully implemented in more than 180 cities, dynamically modifies signal phases in response to actual traffic data, thus improving traffic flow and reducing congestion more effectively than earlier methodologies \cite{aimindSCATS, ndgovSCATS}.

\subsection{Artificial Intelligence in Traffic Signal Control}

The integration of artificial intelligence (AI) into the realm of traffic management has catalyzed substantial advances in efficiency and performance. For example, systems such as SURTRAC utilize a decentralized, schedule-driven approach for intersection control, achieving average reductions in travel time that exceed 25\% in pilot deployments \cite{cmuSURTRAC, jctSURTRAC}. Furthermore, Google's Green Light project exemplifies the application of AI in optimizing traffic signal timings, producing reductions of up to 30\% idle time at red lights accomplished without the need for new hardware installations \cite{googlegreenLight}. This highlights the potential of AI-driven solutions to revolutionize traffic signal control by enabling a more fluid and responsive traffic management system.

\subsection{Reinforcement Learning Approaches}

Reinforcement Learning (RL) has gained prominence as a potent tool for adaptive control of traffic signals. Although single-agent RL methodologies have demonstrated their effectiveness in optimizing individual intersection operations, they often encounter scalability challenges when applied to larger traffic networks \cite{arel2010reinforcement}. To mitigate these limitations, Multi-Agent Reinforcement Learning (MARL) frameworks have been proposed, facilitating decentralized decision-making among multiple agents \cite{li2022multiagent, luo2024multiagent}.

In a significant advancement, a scalable MARL algorithm was introduced that utilizes the Advantage Actor-Critic (A2C) approach, specifically designed for large-scale traffic networks \cite{chu2019scalable}. This method exhibits both robustness and sample efficiency and is effective in both synthetic environments and real-world applications. Building on this foundation, QCOMBO was developed as a MARL algorithm that merges independent and centralized learning strategies \cite{zhang2019qcombo}. This approach achieves competitive performance in a variety of road topologies, thereby enhancing the versatility of MARL applications in traffic control. Additionally, IG-RL was created to use graph-convolutional networks to facilitate the development of transferable adaptive traffic signal control policies that adapt to various network structures \cite{devailly2020igrl}.

Furthering the exploration of cooperative strategies, KS-DDPG was presented as an innovative framework that integrates a knowledge-sharing communication protocol between agents \cite{li2021ksddpg}. This enhancement significantly improves cooperation and control efficiency within large-scale transportation networks. Moreover, the exploration of Mean Field Reinforcement Learning methods has revealed promising results, showcasing improved convergence rates and reduced vehicle loss times compared to traditional deep RL methodologies.

\subsection{Multimodal and Safety-Oriented Traffic Control}

Recent investigations have increasingly turned their attention to multi-modal traffic environments and the imperative of safety optimization in signal control. A noteworthy contribution in this domain is the decentralized MARL approach known as eMARLIN-MM, which was specifically designed to optimize signal timings for both vehicular and transit traffic. This approach has successfully achieved significant reductions in total delays in people's time \cite{emerlin2024}. In a complementary study, Essa and Sayed developed RS-ATSC, an adaptive traffic signal control algorithm that is capable of optimizing intersection safety in real-time, utilizing data derived from connected vehicles \cite{essa2023rsat}.

\subsection{Future Directions}

Comprehensive reviews of the literature have emphasized the transformative potential of integrating AI, IoT, and predictive analytics into adaptive traffic control systems. Various techniques, including fuzzy logic, deep neural networks, and hybrid models that fuse Long Short-Term Memory networks (LSTMs) with Convolutional Neural Networks (CNNs), have been used to analyze complex traffic patterns. These methodologies enable dynamic adjustment of signal timings, thus effectively alleviating congestion and minimizing queueing issues. As the field continues to evolve, ongoing research will likely further refine these approaches, promoting smarter and safer traffic management solutions.

\subsection{Summary}

The evolution of traffic signal control methodologies signifies a critical shift towards the implementation of intelligent and adaptive systems that can respond effectively to real-time traffic dynamics. Multi-Agent Reinforcement Learning (MARL) approaches, in particular, have emerged as promising solutions that offer scalable and decentralized strategies for managing the complexities inherent in urban traffic networks. This research endeavors to build upon these advancements by rigorously comparing the performance of a traditional fixed-time traffic signal controller with that of a MARL-based controller in a simulated multi-intersection environment. The findings of this comparative study aim to contribute valuable insights to the ongoing development of efficient traffic management strategies, ultimately enhancing urban mobility and reducing congestion.

\section{Methodology}

To assess the efficacy of fixed-time control in comparison to Multi-agent Reinforcement Learning (MARL) for optimizing traffic signal management, a traffic simulation environment utilizing the Pygame \cite{pygameDocs} library was developed. This environment enables systematic experimentation across a spectrum of traffic scenarios and control strategies. The source code for this simulation is publicly accessible in the GitHub Repository \cite{githubRepo}.

\subsection{Simulation Environment}

The simulation environment is meticulously designed to replicate a simplified urban traffic network, providing a platform to analyze traffic flow dynamics under various signal control methodologies.

\subsubsection{Simulation Framework}

At the core of the simulation is Pygame \cite{pygameDocs}, a versatile cross-platform suite of Python modules specifically designed for game development. Pygame \cite{pygameDocs} offers extensive capabilities for rendering graphics and managing real-time events. Its inherent flexibility makes it particularly suitable for constructing custom simulations that require dynamic visualizations and interactive user engagement.

\subsubsection{Road Network Configuration}

The simulated road network is structured as a $900 \times 900$ pixel grid, designed to represent a four-way intersection characterized by four traffic lights positioned at the coordinates $(300, 300)$, $(600, 300)$, $(300, 600)$, and $(600, 600)$. Each intersection operates independently, allowing for the implementation and comparative analysis of various traffic signal control strategies.

\begin{figure}[!t]
    \centering
    \includegraphics[width=\linewidth]{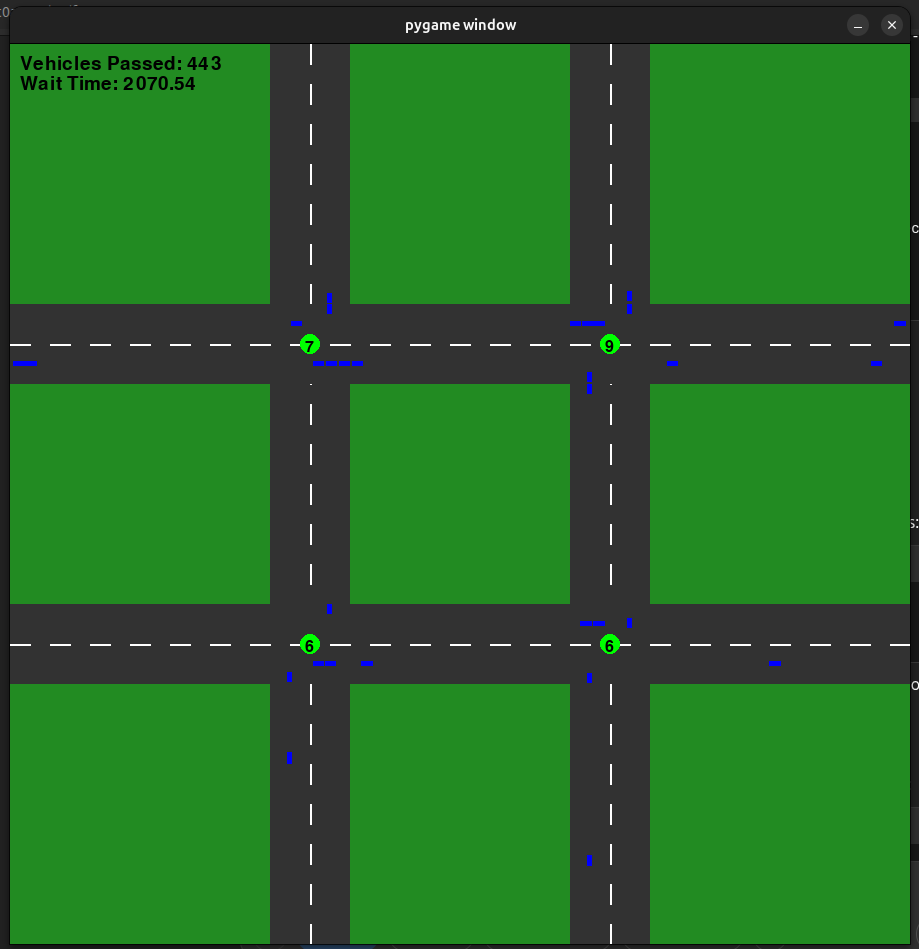}
    \caption{Screenshot of the Pygame-based traffic simulation environment}
    \label{fig:simulation}
\end{figure}

A representative screenshot of the simulation environment is shown in Figure~\ref{fig:simulation}. This figure captures the intricate layout of the road network, including the positioning of traffic signals and vehicles, as well as the vehicle count displays observed during a typical simulation run.

\subsubsection{Traffic Signal Parameters}

Traffic signals function on a predetermined fixed-time cycle, comprising three distinct phases: 
\begin{itemize}
    \item \textbf{Green Phase}: Taking place for $5$ seconds, during this time, vehicles traveling in the horizontal direction are allowed to proceed, while vertical vehicle movement is stopped.
    \item \textbf{Red Phase}: Also spanning $5$ seconds, this phase allows vehicles moving in the vertical direction to advance, while vehicles in the horizontal direction must remain stationary.
    \item \textbf{Yellow Phase}: This $2$-second interval requires that all vehicles, irrespective of their direction, come to a complete stop.
\end{itemize}
The operational behavior of the traffic lights is uniformly applied across all intersections in the fixed-time control simulations. In contrast, in simulations using Multi-Agent Reinforcement Learning (MARL), the signal phases are dynamically adjusted according to the policies developed by the learning agents.

\subsubsection{Vehicle Generation and Movement}

To simulate realistic traffic patterns, vehicles are generated at eight designated entry points at precise intervals of $0.5$ seconds. The specific spawn locations and their corresponding movement directions are delineated as follows:
\begin{itemize}
    \item $(0, 280)$ moving right
    \item $(0, 580)$ moving right
    \item $(900, 320)$ moving left
    \item $(900, 620)$ moving left
    \item $(320, 0)$ moving down
    \item $(620, 0)$ moving down
    \item $(280, 900)$ moving up
    \item $(580, 900)$ moving up
\end{itemize}
Each vehicle adheres to a predetermined trajectory, observing crucial traffic rules such as stopping at red signals and proceeding during green phases. For simplicity and to maintain consistency in the simulation environment, vehicles are modeled to travel at a constant speed and are restricted to linear movement. Furthermore, vehicles are programmed to stop when a vehicle traveling in the same direction comes to a halt immediately ahead of them.

\subsubsection{Traffic Light Vehicle Count Display}

To augment the analytical capabilities of the simulation framework, each traffic light is equipped with a display mechanism that provides a real-time count of approaching vehicles. This feature serves as a crucial tool for monitoring local traffic density at each intersection, thus facilitating a more comprehensive assessment of traffic flow and potential levels of congestion. The implementation of such displays has been shown to improve signal timing optimization, as documented in previous traffic simulation studies.

\subsubsection{Simulation Dynamics}

The simulation operates at a frame rate of \(60\) frames per second (FPS) over a total duration of \(10\) minutes for each experimental run. Throughout each simulation, data are systematically collected on critical performance metrics, specifically:

\begin{itemize}
    \item The total number of vehicles that successfully navigate the intersection and exit
    \item The average wait time experienced by each vehicle
\end{itemize}

These metrics serve as fundamental quantitative indicators to assess the effectiveness of various traffic signal control strategies.

\subsubsection{Relevance and Applications}

Using a Pygame-based simulation facilitates a controlled and adaptable environment for the rigorous testing and comparative analysis of diverse traffic control strategies. This approach is consistent with methodologies utilized in previous studies aimed at modeling and examining traffic systems. The insights derived from such simulations have the potential to significantly enhance the development of more efficient traffic management systems in real-world urban contexts.

\subsection{Traffic Signal Control Strategies}

Effective management of urban traffic flow is essential to optimize transportation systems. This study explores two principal traffic signal control strategies: the traditional fixed-time controller and a Multi-Agent Reinforcement Learning (MARL) based controller. This section provides an in-depth examination of the fixed-time control approach utilized in the simulation.

\subsection{Fixed-Time Controller}

The fixed-time control strategy is characterized by its operation on an established schedule in which traffic signal phases transition through fixed durations without regard for real-time traffic conditions. This method is commonly adopted in traffic management frameworks due to its inherent simplicity and ease of implementation, particularly in contexts where traffic patterns exhibit relative consistency or where infrastructural limitations preclude the adoption of adaptive control mechanisms \cite{web26, nacto2025fixed, study2025}.

\subsubsection{Determination of Fixed Timings}

In the simulation, the fixed timings were selected primarily to enable efficient data collection within practical simulation time constraints. Although cycle lengths and phase durations do not strictly follow established traffic engineering optimization principles, they were designed to represent plausible urban intersection scenarios.

The total cycle length was calibrated to 24 seconds (12 seconds for North-South and 12 seconds for East-West), incorporating the following phases:

\begin{itemize}
    \item \textbf{Green Phase}: 5 seconds
    \item \textbf{Yellow Phase}: 2 seconds
    \item \textbf{Red Phase}: 5 seconds
\end{itemize}

This sequence facilitates the alternating green phases for perpendicular traffic flows.

\subsubsection{Implementation in Simulation}

In the simulation environment, each traffic signal operates on a predetermined fixed-time schedule, operating independently of adjacent signals. The controller systematically cycles through predefined phases, updating the status of the traffic lights at each interval. Vehicles react to the current state of the traffic signal at their respective intersections, advancing during the green phases while stopping during the red and yellow phases. 

This methodology establishes a fundamental baseline for assessing the efficacy of more sophisticated control strategies, such as the Multi-Agent Reinforcement Learning (MARL)-based controller. By providing consistent and predictable traffic signal behavior, the fixed-time approach enables a clear framework against which the improvements offered by advanced methodologies can be evaluated.

\subsection{Multi-Agent Deep Q-Network Architecture}

The approach uses a multi-agent reinforcement learning framework where each traffic light is controlled by an independent Deep Q-Network (DQN) agent. Each agent observes a global state representation and selects actions to optimize local traffic flow, while implicitly coordinating with other agents through shared state information \cite{luo2024multiagent, academia109014103}. This decentralized control mechanism improves scalability and robustness in large traffic networks, since each agent independently learns and adapts to traffic conditions, yet maintains overall network efficiency through shared information \cite{web9}.

\subsubsection{Neural Network Architecture}

Each DQN agent utilizes a fully connected feedforward neural network with the following architecture:
\begin{itemize}
    \item Input layer: $n \times 20$ dimensions, where $n$ is the number of traffic lights
    \item Hidden layer 1: 128 neurons with ReLU activation
    \item Hidden layer 2: 64 neurons with ReLU activation
    \item Output layer: $|\mathcal{A}|$ neurons (corresponding to available light states)
\end{itemize}

\begin{figure}[!t]
    \centering
    \includegraphics[width=\linewidth]{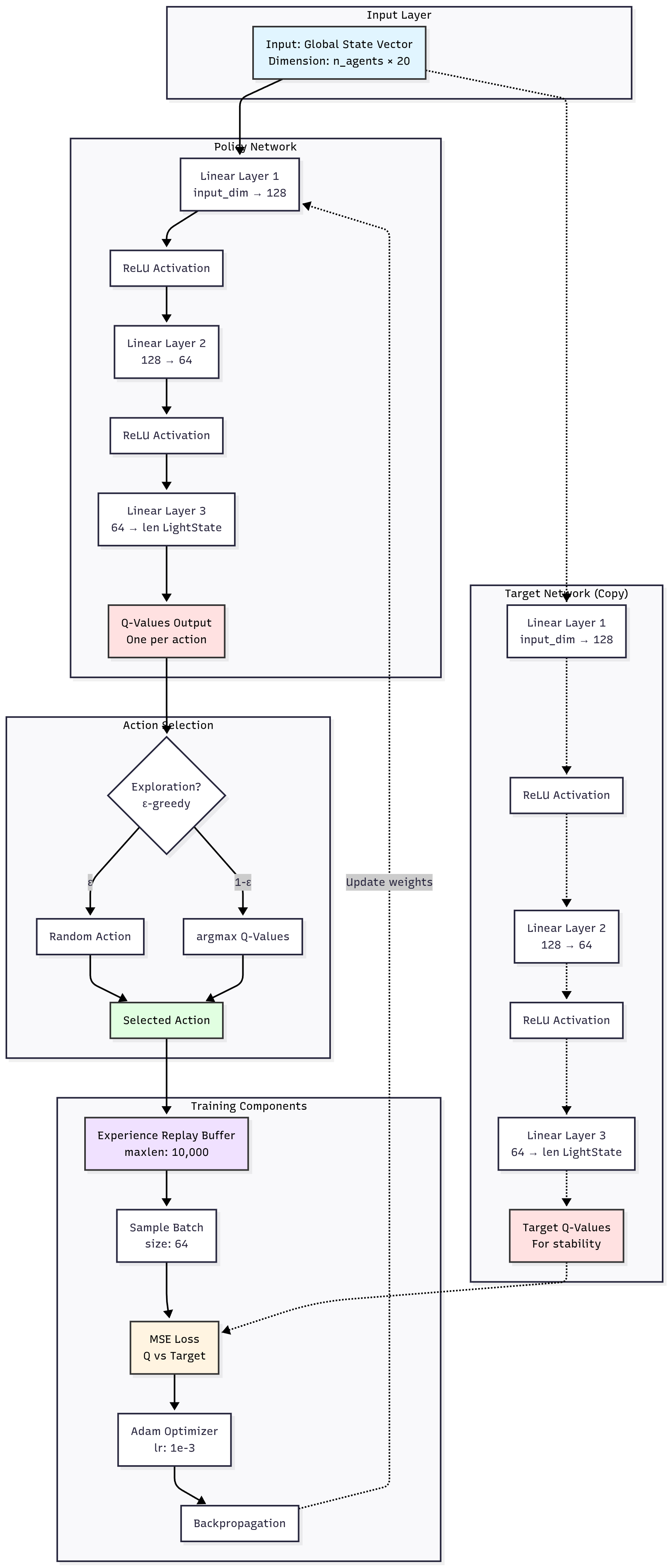}
    \caption{Deep Q-Network Architecture}
    \label{fig:dqn_architecture}
\end{figure}

The network approximates the Q-function $Q(s, a; \theta)$, which estimates the expected cumulative reward for taking action $a$ in state $s$ with parameters $\theta$.

\subsubsection{State Representation}

The global state vector concatenates local observations from all traffic lights, where each light contributes 20 features:
\begin{itemize}
    \item \textbf{Queue lengths} (4 features): Number of vehicles approaching from each cardinal direction (N, S, E, W)
    \item \textbf{Average distances} (4 features): Mean distance to the traffic light for vehicles in each direction
    \item \textbf{Movement ratios} (4 features): Proportion of moving vehicles in each approaching direction
    \item \textbf{Spatial features} (8 features): Average relative positions ($\Delta x$, $\Delta y$) of vehicles in each direction
\end{itemize}

\begin{figure}[!t]
    \centering
    \includegraphics[width=\linewidth]{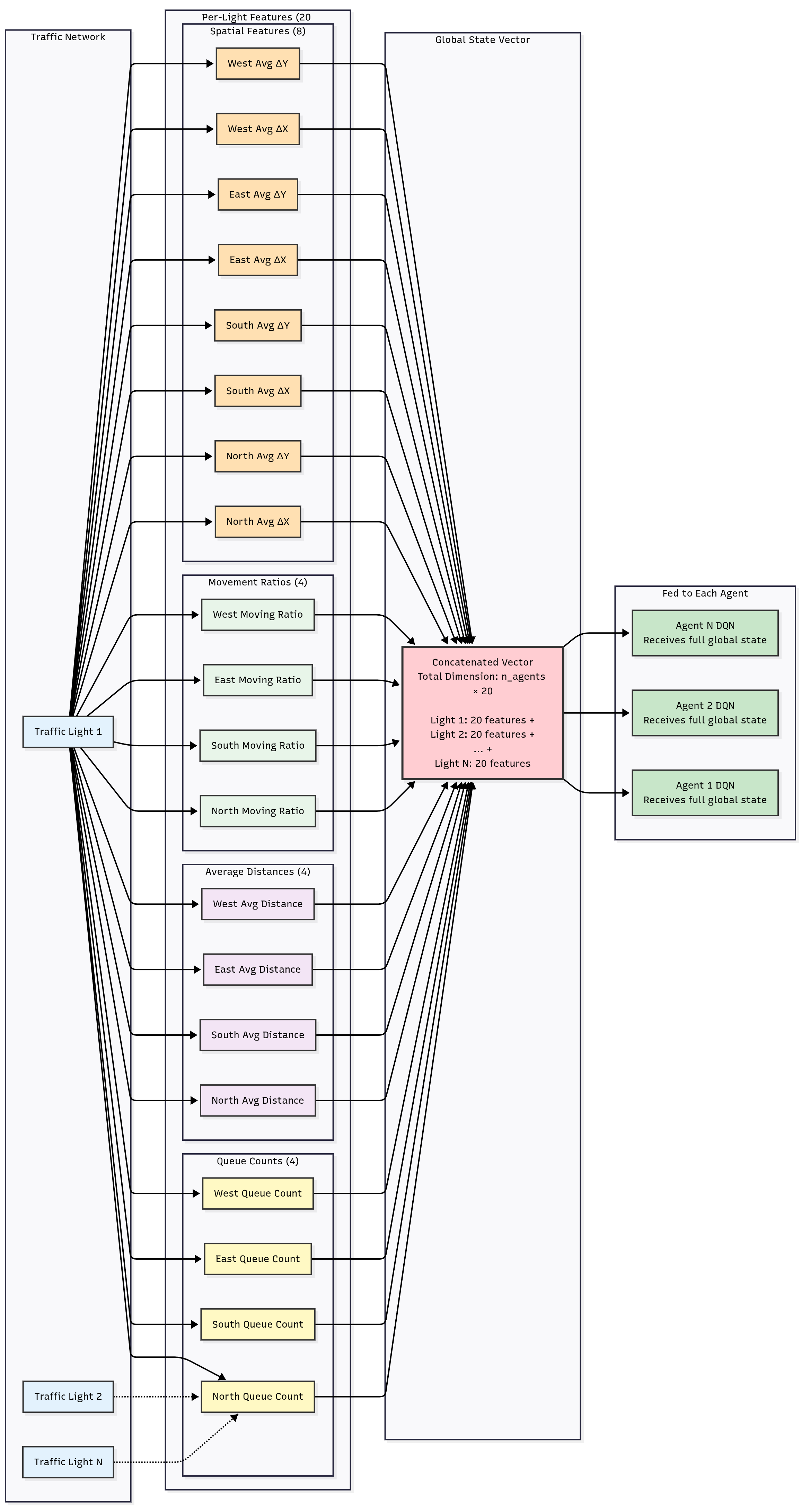}
    \caption{State Representation}
    \label{fig:state_representation}
\end{figure}

This global state representation enables implicit coordination between agents by providing visibility into traffic conditions at all intersections.

\subsection{Reinforcement Learning Framework}

\subsubsection{Action Space and Constraints}

Each agent selects from the discrete action space $\mathcal{A} = \{\text{GREEN}, \text{RED}, \text{YELLOW}\}$ corresponding to the available traffic light states. To ensure realistic traffic signal behavior, temporal constraints were enforced:
\begin{itemize}
    \item \textbf{Minimum duration}: $t_{\min} = 1.0$ seconds - prevents rapid switching
    \item \textbf{Maximum duration}: $t_{\max} = 10.0$ seconds - ensures fairness across directions
\end{itemize}

When $t < t_{\min}$, the previous action is kept. When $t \geq t_{\max}$, the agent is forced to switch to the next-best action according to the current Q-values.

\subsubsection{Reward Function}

The reward function balances throughput maximization with congestion minimization:
\begin{equation}
r_t = \text{moved}_t - 0.1 \cdot \text{queue}_t - 0.2 \cdot \text{stopped}_t
\end{equation}

where:
\begin{itemize}
    \item $\text{moved}_t$: Number of vehicles that pass through the intersection in favorable directions
    \item $\text{queue}_t$: Total queue length in all directions
    \item $\text{stopped}_t$: Number of stopped vehicles in all directions
\end{itemize}

This reward structure incentivizes maximizing vehicle throughput while penalizing queue buildup and vehicle stoppages \cite{iet2024reward, ceur2022reward, aston2020reward}.

\subsection{Training Algorithm}

\subsubsection{Experience Replay}

Experience replay is a crucial technique used to stabilize the training of deep reinforcement learning models for traffic signal control. Each agent maintains a replay buffer $\mathcal{D}_i$ with a capacity of 10,000 transitions, which stores tuples $(s_t, a_t, r_t, s_{t+1}, d_t)$, where $d_t$ indicates whether the episode has terminated \cite{gao2017adaptive, liang2018deep, swapno2024traffic}.

This buffer allows the model to learn from a diverse set of past experiences, breaking the correlation between sequential data, thus improving training stability and convergence \cite{gao2017adaptive,swapno2024traffic}. The stored transitions are sampled randomly during training updates, which helps mitigate the risk of overfitting to recent experiences and smooths the learning process \cite{liang2018deep}.

\subsubsection{Q-Learning Update}

At each time step, after all agents execute their actions and observe rewards, the following update were performed for each agent $i$:

\begin{enumerate}
    \item Sample a mini-batch of $B = 64$ transitions from $\mathcal{D}_i$
    \item Compute current Q-values: $Q(s_t, a_t; \theta_i)$
    \item Compute target Q-values using the target network:
    \begin{equation}
    y_t = r_t + \gamma \max_{a'} Q(s_{t+1}, a'; \theta_i^-)
    \end{equation}
    where $\theta_i^-$ are the target network parameters and $\gamma = 0.99$ is the discount factor
    \item Minimize the loss function:
    \begin{equation}
    \mathcal{L}(\theta_i) = \mathbb{E}_{(s,a,r,s') \sim \mathcal{D}_i}\Big[\Big(y_t - Q(s_t, a_t; \theta_i)\Big)^2\Big]
    \end{equation}
    \item Update network parameters using Adam optimizer with learning rate $\alpha = 10^{-3}$
\end{enumerate}

\subsubsection{Target Network and Exploration}

To improve training stability, a target network $Q(s, a; \theta^-)$ is updated every 100 time steps by copying the weights from the main network: $\theta^- \leftarrow \theta$.

$\epsilon$-greedy exploration with decaying epsilon was used:
\begin{equation}
a_t = \begin{cases}
\text{random action} & \text{with probability } \epsilon_t \\
\arg\max_{a} Q(s_t, a; \theta) & \text{with probability } 1 - \epsilon_t
\end{cases}
\end{equation}

where $\epsilon_t$ decays linearly from $\epsilon_0 = 1.0$ to $\epsilon_{\min} = 0.05$ with decay rate $\lambda = 10^{-4}$ per update step \cite{gao2017adaptive, liang2018deep, swapno2024traffic}.

\subsection{Training Procedure}

The complete training procedure operates as follows:

\begin{algorithm}
\caption{Multi-Agent Traffic Light Training}
\begin{algorithmic}[1]
\STATE Initialize DQN agents $\{Q_i\}_{i=1}^n$ with random weights
\STATE Initialize target networks $\{Q_i^-\}_{i=1}^n$ with $\theta_i^- \leftarrow \theta_i$
\STATE Initialize replay buffers $\{\mathcal{D}_i\}_{i=1}^n$
\FOR{each timestep $t$}
    \STATE Observe global state $s_t$
    \FOR{each agent $i$}
        \STATE Select action $a_i^t$ using $\epsilon$-greedy policy
        \STATE Apply temporal constraints
        \STATE Execute action and update traffic light state
    \ENDFOR
    \STATE Observe next global state $s_{t+1}$
    \FOR{each agent $i$}
        \STATE Calculate reward $r_i^t$
        \STATE Store transition $(s_t, a_i^t, r_i^t, s_{t+1})$ in $\mathcal{D}_i$
        \STATE Sample mini-batch and perform Q-learning update
    \ENDFOR
    \IF{$t \mod 100 = 0$}
        \STATE Update all target networks: $\theta_i^- \leftarrow \theta_i$ for all $i$
    \ENDIF
\ENDFOR
\end{algorithmic}
\end{algorithm}

\begin{figure}[!t]
    \centering
    \includegraphics[width=\linewidth]{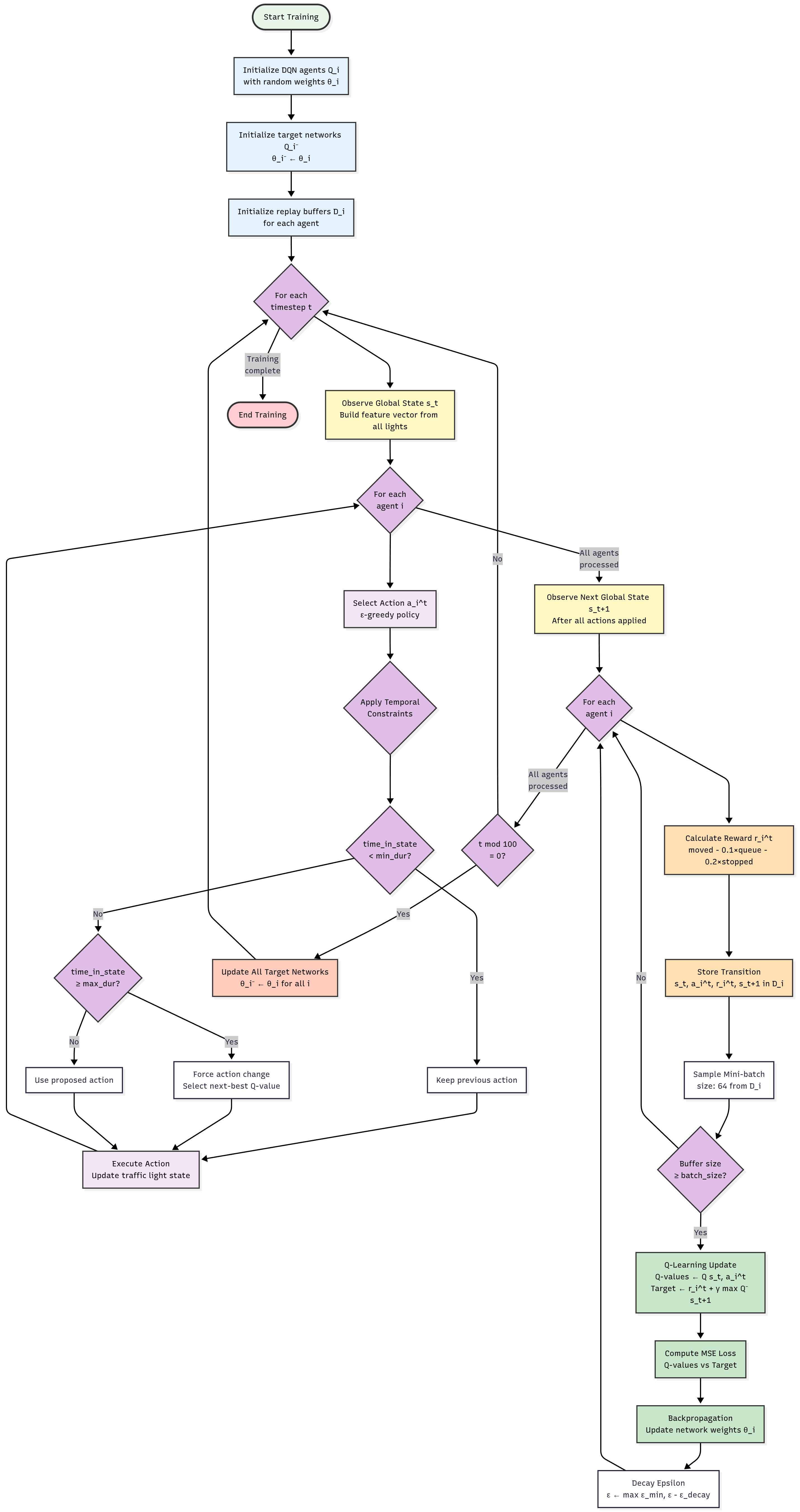}
    \caption{Full Training Procedure}
    \label{fig:training_procedure}
\end{figure}

This decentralized training approach allows each agent to learn independently while coordinating implicitly through the shared global state representation, enabling scalable multi-agent traffic control.

\subsection{Experimental Design}

\subsubsection{Simulation Framework}

To rigorously evaluate the performance of the Fixed-Time (FT) controller compared to the Multi-Agent Reinforcement Learning (MARL) controller, a discrete-event traffic simulation model was created. Within this framework, vehicles are randomly generated at one of eight predefined locations at intervals of 0.5 seconds. All other elements of the simulation—including vehicle dynamics and traffic signal operations—are deterministic. This methodological structure ensures that the sole source of stochasticity lies in the vehicle spawning process, thereby permitting a controlled and valid comparison between the two controller types.

\subsubsection{Handling Randomness and Ensuring Fairness}

The inherent randomness associated with vehicle spawning requires the execution of multiple simulation runs to achieve statistically significant findings. Relying on a singular simulation run risks yielding misleading conclusions, potentially skewed by outlier effects. 20 independent simulation runs were performed for each type of controller, each using a distinct random seed to promote variability in vehicle spawning patterns.

This methodology is in line with established best practices in traffic simulation, which advocate running multiple simulations to adequately capture the intrinsic variability of stochastic systems \cite{ptv2023traffic, txdot2013simulation, wsdot2021protocol}. By averaging the results across these runs, the influence of random fluctuations is mitigated, thereby facilitating a fair and equitable comparison between the fixed-time and MARL controllers.

\subsubsection{Justification for Experimental Design}

The decision to implement 20 simulation runs per controller is strongly supported in the literature, which highlights the critical importance of multiple iterations in stochastic simulations to generate reliable and valid results. Furthermore, isolating the randomness solely from the vehicle spawning process ensures that any observed performance differentials are attributable directly to the controllers themselves, devoid of interference from extraneous variables.

By adhering to such rigorous methodological principles, the experimental design provided a robust and fair comparison between the fixed-time and MARL controllers \cite{ptv2023traffic, txdot2013simulation, wsdot2021protocol}.

\subsection{Performance Metrics}

In the investigation of traffic signal control strategies, two primary performance metrics served as critical indicators of efficacy:

\begin{itemize}
    \item \textbf{Average Vehicle Delay:} This metric calculates the mean duration that vehicles are required to wait at intersections, thereby providing a direct reflection of traffic flow efficiency and the effectiveness of signal timing strategies. The Highway Capacity Manual (HCM) identifies the average vehicle delay as a fundamental metric for assessing the Level of Service (LOS) at signalized intersections, highlighting its pivotal role in evaluating traffic performance \cite{fhwa2008hcm, ctrutexas2019, sciencedirect2025}.
    
    \item \textbf{Throughput:} Throughput is defined as the total number of vehicles that successfully traverse the traffic network in a specified time. This metric serves as a quantitative indicator of both the capacity and efficiency of the traffic system. High throughput values are indicative of improved traffic movement and reduced levels of congestion \cite{rosap2018, fhwa2008hcm}.
\end{itemize}

Together, these metrics provide a comprehensive framework for assessing the performance of traffic signals. Although average vehicle delay emphasizes the temporal aspect of traffic efficiency, throughput addresses the volumetric capacity of the network.

\section{Results}

The performance of the fixed-time traffic controller and the Multi-Agent Reinforcement Learning (MARL) controller was evaluated through 20 simulation runs for each approach. The simulations measured two key performance indicators: the number of vehicles that passed through the intersection and the average waiting time experienced by the vehicles. Tables~\ref{tab:fixed_results} and~\ref{tab:marl_results} present the complete results of all simulation runs.

\subsection{Statistical Methodology}

To determine whether the observed differences between the fixed-time and MARL controllers are statistically significant, hypothesis testing was performed for both performance metrics.

\subsubsection{Hypotheses}

For each metric, null and alternative hypotheses were formulated:

\textbf{For Vehicles Passed:}
\begin{itemize}
    \item $H_0$: $\mu_{MARL} = \mu_{Fixed}$ (no difference in vehicles passed)
    \item $H_1$: $\mu_{MARL} > \mu_{Fixed}$ (MARL passes more vehicles)
\end{itemize}

\textbf{For Wait Time:}
\begin{itemize}
    \item $H_0$: $\mu_{MARL} = \mu_{Fixed}$ (no difference in wait time)
    \item $H_1$: $\mu_{MARL} < \mu_{Fixed}$ (MARL has shorter wait times)
\end{itemize}

\subsubsection{Test Selection}

Given that there are two independent samples of equal size ($n = 20$ for each controller) and comparing their means, the independent samples t-test was used. Since there is a directional hypothesis (MARL is expected to perform better), one-tailed tests were used. This statistical approach is appropriate when data distributions are approximately normal and variances are equal, as confirmed by preliminary diagnostics \cite{vidali2021comparative, nchrp1991traffic}.

The t-test statistic is calculated as follows:
\begin{equation}
t = \frac{\bar{x}_1 - \bar{x}_2}{s_p\sqrt{\frac{2}{n}}}
\end{equation}

where $\bar{x}_1$ and $\bar{x}_2$ are the sample means, $n$ is the sample size, and $s_p$ is the pooled standard deviation:
\begin{equation}
s_p = \sqrt{\frac{(n_1-1)s_1^2 + (n_2-1)s_2^2}{n_1+n_2-2}}
\end{equation}

The assumptions for the t-test are the following:
\begin{enumerate}
    \item Independence: Each simulation run is independent
    \item Normality: The data should be approximately normally distributed (verified using Shapiro-Wilk test)
    \item Equal variances: The variances of both groups should be approximately equal (verified using Levene's test)
\end{enumerate}

If the equal variance assumption is violated, Welch's t-test should be used, which does not assume equal variances. The significance level was set at $\alpha = 0.05$ \cite{nchrp1991traffic, fhwa2008tft, mautcpennstate2008}.

\subsection{Fixed-Time Controller Results}

Table~\ref{tab:fixed_results} shows the results of 20 simulation runs using the fixed-time traffic controller.

\begin{table}[ht]
\centering
\caption{Fixed-Time Controller Simulation Results}
\label{tab:fixed_results}
\begin{tabular}{ccc}
\hline
\textbf{Run} & \textbf{Vehicles Passed} & \textbf{Wait Time (s)} \\
\hline
1 & 1146 & 5283.23 \\
2 & 1148 & 5193.31 \\
3 & 1146 & 5262.83 \\
4 & 1146 & 5081.75 \\
5 & 1144 & 5250.06 \\
6 & 1146 & 5327.18 \\
7 & 1146 & 5228.23 \\
8 & 1146 & 5347.56 \\
9 & 1146 & 5310.16 \\
10 & 1149 & 5249.45 \\
11 & 1148 & 5280.27 \\
12 & 1148 & 5382.48 \\
13 & 1145 & 5244.82 \\
14 & 1143 & 5227.23 \\
15 & 1147 & 5374.63 \\
16 & 1145 & 5370.44 \\
17 & 1146 & 5150.83 \\
18 & 1147 & 5364.79 \\
19 & 1149 & 5142.19 \\
20 & 1147 & 5205.03 \\
\hline
\end{tabular}
\end{table}

\subsection{MARL Controller Results}

Table~\ref{tab:marl_results} shows the results of 20 simulation runs using the MARL traffic controller.

\begin{table}[ht]
\centering
\caption{MARL Controller Simulation Results}
\label{tab:marl_results}
\begin{tabular}{ccc}
\hline
\textbf{Run} & \textbf{Vehicles Passed} & \textbf{Wait Time (s)} \\
\hline
1 & 1154 & 1149.46 \\
2 & 1155 & 1104.94 \\
3 & 1154 & 1156.51 \\
4 & 1153 & 1111.07 \\
5 & 1155 & 1107.69 \\
6 & 1152 & 1128.64 \\
7 & 1152 & 1208.07 \\
8 & 1153 & 1166.85 \\
9 & 1150 & 1129.98 \\
10 & 1153 & 1135.63 \\
11 & 1154 & 1154.45 \\
12 & 1152 & 1173.42 \\
13 & 1152 & 1145.04 \\
14 & 1152 & 1115.58 \\
15 & 1155 & 1174.28 \\
16 & 1155 & 1148.37 \\
17 & 1153 & 1122.54 \\
18 & 1153 & 1137.87 \\
19 & 1153 & 1168.26 \\
20 & 1153 & 1156.70 \\
\hline
\end{tabular}
\end{table}

\subsection{Descriptive Statistics}

Table~\ref{tab:descriptive_stats} presents the descriptive statistics for both controllers in both performance metrics.

\begin{table}[ht]
\centering
\caption{Descriptive Statistics Comparison}
\label{tab:descriptive_stats}
\begin{tabular}{lcccc}
\hline
\textbf{Metric} & \multicolumn{2}{c}{\textbf{Vehicles Passed}} & \multicolumn{2}{c}{\textbf{Wait Time (s)}} \\
\cline{2-5}
 & Fixed-Time & MARL & Fixed-Time & MARL \\
\hline
Mean & 1146.40 & 1153.15 & 5263.82 & 1144.77 \\
Std. Deviation & 1.54 & 1.31 & 84.06 & 26.36 \\
Variance & 2.36 & 1.71 & 7065.51 & 694.74 \\
Minimum & 1143.00 & 1150.00 & 5081.75 & 1104.94 \\
Maximum & 1149.00 & 1155.00 & 5382.48 & 1208.07 \\
Median & 1146.00 & 1153.00 & 5256.45 & 1146.71 \\
\hline
\end{tabular}
\end{table}

\subsection{Statistical Analysis}

\subsubsection{Vehicles Passed Analysis}

 Before hypothesis testing, assumptions for the t-test were verified. The Shapiro-Wilk normality test indicated that both datasets were normally distributed (Fixed-Time: $W = 0.939$, $p = 0.225$; MARL: $W = 0.906$, $p = 0.053$). Levene's test confirmed equal variances between groups ($F = 0.221$, $p = 0.641$). Therefore, the test proceeded with the standard independent sample t-test.

The results showed a statistically significant difference in vehicle throughput:
\begin{itemize}
    \item MARL mean: $\mu_{MARL} = 1153.15$ vehicles
    \item Fixed-Time mean: $\mu_{Fixed} = 1146.40$ vehicles
    \item Difference: 6.75 vehicles (0.59\% improvement)
    \item Test statistic: $t(38) = 14.96$
    \item $p$-value: $p = 8.20 \times 10^{-18}$
    \item Cohen's $d = 4.73$ (large effect size)
\end{itemize}

Since $p < 0.05$, $H_0$ was rejected, and it was concluded that the MARL controller passes significantly more vehicles than the fixed-time controller.

\subsubsection{Wait Time Analysis}

The Shapiro-Wilk test confirmed normality for both datasets (Fixed-Time: $W = 0.960$, $p = 0.537$; MARL: $W = 0.966$, $p = 0.679$). However, Levene's test revealed unequal variances ($F = 15.43$, $p = 3.50 \times 10^{-4}$), necessitating the use of Welch's t-test.

The results demonstrated a highly significant reduction in wait time:
\begin{itemize}
    \item MARL mean: $\mu_{MARL} = 1144.77$ seconds
    \item Fixed-Time mean: $\mu_{Fixed} = 5263.82$ seconds
    \item Difference: 4119.06 seconds (78.25\% reduction)
    \item Test statistic: $t(22.70) = -209.11$
    \item $p$-value: $p = 4.30 \times 10^{-39}$
    \item Cohen's $d = -66.13$ (large effect size)
\end{itemize}

Since $p < 0.05$, $H_0$ was rejected and it was concluded that the MARL controller has a significantly shorter waiting time than the fixed-time controller.

\subsection{Summary}

Statistical analysis provides strong evidence that the MARL controller outperforms the fixed-time controller on both performance metrics. The MARL controller achieved the following:

\begin{enumerate}
    \item A statistically significant 0.59\% increase in vehicle throughput ($p = 8.20 \times 10^{-18}$)
    \item A statistically significant 78.25\% reduction in average wait time ($p = 4.30 \times 10^{-39}$)
\end{enumerate}

Both improvements have large effect sizes (Cohen's $d = 4.73$ and $-66.13$ respectively), indicating that the differences are not only statistically significant, but also practically meaningful. The extremely small p-values provide overwhelming evidence against the null hypotheses, strongly supporting the adoption of MARL for traffic signal control.

\section{Discussion}

This research investigated the effectiveness of Multi-Agent Reinforcement Learning (MARL) for traffic signal control compared to traditional fixed-time controllers. Through 20 independent simulation runs for each control strategy, performance was evaluated in two critical metrics: vehicle throughput and average wait time.

\subsection{Key Findings}

The experimental results provide compelling evidence for the superiority of MARL-based traffic control. Statistical analysis revealed that the MARL controller achieved statistically significant improvements in both performance metrics:

\begin{itemize}
    \item \textbf{Vehicle Throughput}: The MARL controller increased vehicle throughput by 0.59\% (from 1146.40 to 1153.15 vehicles per interval), with an extremely significant p-value of $8.20 \times 10^{-18}$ and a large effect size (Cohen's $d = 4.73$).
    
    \item \textbf{Wait Time Reduction}: The MARL controller reduced the average wait time by 78.25\% (from 5263.82 to 1144.77 seconds), demonstrating a dramatic improvement in traffic flow efficiency with a p-value of $4.30 \times 10^{-39}$ and an exceptionally large effect size (Cohen's $d = -66.13$).
\end{itemize}

The magnitude of the reduction in waiting time is particularly significant from a practical point of view. A reduction of more than 4000 seconds in the average waiting time translates to substantially improved traffic flow, reduced fuel consumption, lower emissions, and increased driver satisfaction. Although the improvement in vehicle throughput appears modest in percentage terms, the large effect size and extremely small p-value indicate that this improvement is consistent and reliable in all simulation runs.

\subsection{Implications}

These findings have important implications for urban traffic management:

\begin{itemize}
    \item \textbf{Adaptive Control Advantage}: The MARL controller's ability to adapt signal timing based on real-time traffic conditions clearly outperforms the rigid, predetermined timing of fixed-time controllers. This adaptive capability allows the system to respond dynamically to traffic patterns, resulting in more efficient intersection management.
    
    \item \textbf{Multi-Agent Coordination}: The multi-agent framework used in this research demonstrates that reinforcement learning approaches can effectively coordinate traffic signals across multiple intersections in complex urban environments. The consistent performance across all simulation runs suggests that the learned policies are robust and generalizable.
    
    \item \textbf{Practical Deployment Potential}: The substantial improvements observed, particularly in waiting time reduction, provide strong justification for investing in MARL-based traffic control systems. The 78.25\% reduction in wait times could translate to significant economic and environmental benefits in real-world deployments.
\end{itemize}

\section{Limitations}

Although the results are highly promising, several limitations should be acknowledged:

\begin{itemize}
    \item \textbf{Simulation Environment}: The experiments were conducted in a simulated environment. Real-world deployment would need to account for additional factors, such as pedestrian crossings, emergency vehicles, adverse weather conditions, and sensor noise that can affect system performance. The behavior of road users in reality may differ from simulated agents, introducing unpredictability that could impact controller effectiveness.
    
    \item \textbf{Traffic Pattern Coverage}: The simulations evaluated the performance under specific traffic conditions over a limited period of time. The generalizability of these results to different traffic patterns, including peak hours, off-peak periods, special events, and varying traffic compositions (e.g., different proportions of vehicles, busses, and emergency vehicles) remains to be validated. Seasonal variations and changes in long-term traffic patterns were not considered in this study.
    
    \item \textbf{Computational Requirements}: The computational requirements for training and deploying MARL controllers in large-scale urban networks have not been thoroughly evaluated. The reinforcement learning training phase can be computationally intensive and the feasibility of real-time inference in resource-constrained embedded systems requires further investigation.
    
    \item \textbf{Comparative Analysis}: This study did not compare MARL with other adaptive control methods such as actuated control, genetic algorithms, or fuzzy logic controllers, limiting the ability to assess relative performance among modern traffic control approaches.
    
    \item \textbf{Network Scalability}: The performance of MARL controllers across different network topologies, varying numbers of intersections, and diverse infrastructure configurations requires further investigation to establish scalability characteristics and generalization capabilities.
\end{itemize}

\section{Future Work}

Several avenues for future research emerge from this study:

\begin{itemize}
    \item \textbf{Diverse Traffic Scenarios}: Evaluate the performance of the MARL controller in a wider range of traffic scenarios, including various traffic densities, time-of-day patterns, seasonal variations, and special event conditions to establish the robustness and generalizability of the approach \cite{li2022multiagent}.
    
    \item \textbf{Transfer Learning}: Investigate the transferability of learned policies to different network configurations and traffic patterns. Research into transfer learning and domain adaptation techniques could enable trained controllers to be deployed more efficiently at different locations with minimal retraining \cite{chu2019scalable, zhang2019qcombo}.
    
    \item \textbf{Comparative Benchmarking}: Conduct comprehensive comparisons with other adaptive control methods. Benchmark MARL against actuated control, adaptive traffic signal control systems (SCATS, SCOOT), genetic algorithms, and fuzzy logic controllers to provide a complete picture of its relative advantages and disadvantages \cite{abdulhai2024emarlin, li2021ksddpg}.
    
    \item \textbf{Real-World Validation}: Conduct pilot deployments in real-world settings to validate simulation results. Field trials would reveal practical challenges related to sensor integration, communication infrastructure, system reliability, and maintenance requirements that cannot be fully captured in simulation \cite{mousavinejad2023multi, essa2023rsat}.
    
    \item \textbf{Hybrid Control Approaches}: Explore hybrid approaches that combine MARL with traditional control methods to enhance robustness. Incorporating rule-based safety constraints or fallback mechanisms could improve system reliability and facilitate regulatory approval for real-world deployment \cite{abdulhai2024emarlin}.
    
    \item \textbf{Connected Vehicle Integration}: Investigate the integration of connected vehicle data and V2X communication to further improve MARL controller performance. Access to vehicle trajectory predictions, driver intentions, and cooperative awareness messages could enable more proactive and efficient signal control strategies \cite{essa2023rsat, abdulhai2024emarlin}.
\end{itemize}

\section{Conclusion}

This study demonstrates that Multi-Agent Reinforcement Learning represents a significant advancement in traffic signal control technology, achieving a statistically significant 0.59\% increase in vehicle throughput and a remarkable 78.25\% reduction in average wait time compared to fixed-time controllers. The extremely small p-values ($8.20 \times 10^{-18}$ and $4.30 \times 10^{-39}$) and the large effect sizes (Cohen's $d = 4.73$ and $-66.13$) provide overwhelming evidence for the practical value of MARL-based traffic management systems. Although limitations related to simulation environments and the need for real-world validation remain, the dramatic improvements in traffic flow efficiency observed in this research strongly support the continued development and deployment of reinforcement learning-based traffic control systems. As urban populations continue to grow and traffic congestion intensifies, MARL controllers offer a promising path toward more efficient, sustainable, and livable cities.

\bibliographystyle{IEEEtran}
\bibliography{references}

\end{document}